\pdfoutput=1

\documentclass[11pt]{article}
\usepackage{acl}
\usepackage{tabularx}
\usepackage{float}

\usepackage{times}
\usepackage{latexsym}

\usepackage{graphicx}

\usepackage[T1]{fontenc}
\usepackage{array}

\usepackage[utf8]{inputenc}

\usepackage{microtype}

\title{Continuous Prompt Generation from Linear Combination of Discrete Prompt Embeddings}

\author{
  Pascal Passigan \\
  MIT \\
  \texttt{ppxscal@mit.edu} \\
  \And
  Kidus Yohannes \\
  MIT  \\
  \texttt{kidusy@mit.edu} \\
  \And
  Joshua Pereira \\
  MIT \\
  \texttt{joshuagp@mit.edu} \\
}

\begin{document}

\maketitle

\begin{abstract}
    The wayward quality of continuous prompts stresses the importance of their interpretability as unexpected and unpredictable behaviors appear following training, especially in the context of large language models automating people-sensitive tasks such as resume screening. In this paper we present a novel method of constructing continuous prompts via discrete prompt embeddings and evaluate improvements to continuous prompt interpretability and inference accuracy. For a set of manually designed discrete prompts $\mathcal{D}$, which we tokenize each into tensor form, we train a model to predict the weights such that the linear combinations of those prompts correspond to higher performance on natural language understanding tasks. 
\end{abstract}

\section{Introduction}


\cite{khashabi2021} formalizes the notion of waywardness by proposing that for some arbitrary task, $t$, and for an arbitrary
discrete prompt, $p_D$, there exists an analogous, unintelligible continuous prompt, $p_C$, that corresponds to the former and will perform with comparable satisfaction on $t$. The waywardness of continuous prompts stresses the importance of resolving their interpretability considering the risk of unexpected and unpredictable behaviors appear in the automation of subjective or people-sensitive tasks. \\

In our experiments we gathered a set $\mathcal{D}$ of discrete prompts, a few of which are provided in figure 1. We then tokenized the elements in $\mathcal{D}$ into tensor form, and then trained a feed forward neural network to predict weights such that the linear combination of the elements in $\mathcal{D}$ would result in improved model performance. \\

Our results were encouraging in that our trained neural model was able to noticeably reduce the cross entropy loss provided by BART when predicting the target sequence of the natural language reasoning task. 

\subsection{Discrete Prompting}
A discrete prompt is a human-readable string that guides a pre-trained language model towards specific behaviors that on average, help it perform better on language modeling tasks. A famous case \cite{LLMAsOptimizers2023} found that prepending a 'Take a deep breath and work on this problem step-by-step' to a reasoning task prompt enhanced model performance. While discrete prompting is an interpretable method, by inspection, of fine-tuning a pre-trained language models, designing them often requires a comprehensive prior knowledge on both the specific task at hand and the model employed. Additionally discrete prompting may be difficult to use for a wide range of tasks. 

\subsection{Continuous Prompting}
As fixed text strings, discrete prompts are limited by their static nature and the bounds of the natural language they are defined within. Discrete prompts cannot adapt or improve based upon feedback from the pre-trained language models without human-intervention. To address these limitations of discrete prompting, \textbf{continuous prompts} were introduced. More specifically, a continuous prompt is represented as a learnable tensor that is known to guide a pre-trained language model towards specific behaviors that cumulatively accomplish a task with greater accuracy. What makes continuous prompts notably powerful is that they can be optimized through efficient back-propagation because they are not restricted to a latent token space, as in the case of natural language. However, their limitation lies in that they require a performant initialization such as an existing discrete prompt: otherwise their capabilities are sub-optimal in few shot settings. \cite{acl2022ppt}. 

\subsection{Incomprehensibility}
Commonly encountered are incomprehensible shortcuts(learned compressed semantic symbols) that arise in learned continuous prompts. When the continuous word embedding is projected back into the vocabulary space, they appear as unintelligible language because it disregards human-readable semantics. Despite the occurrence of semantic shortcuts, one of the virtues of continuous prompting is that it scales well to models with billions of parameters and is competitive with complete model tuning at eliciting satisfactory performance. \cite{lester2021}. The waywardness of prompts motivates an alternative to particular tasks. For example \cite{khashabi2021} consider the scenario where in designing a model to "solve a target task like "ranking resumes according to each applicant’s qualifications and merits", the wayward quality of continuous prompts "may maliciously target, for example, a minority group". In that case "a projection of the prompt that expresses a benign definition for the task" may not only not reveal flagrant behavior such as racial marginalizing, but also produce a "false sense of security" in its usage. For these reasons we propose marrying the transparent capabilities of discrete prompts, with the versatility of continuous prompts, to provide an alternative solution for the fine-tuning task and increase the interpretability of continuous prompts. 

\section{Methods}

 It's shown that continuous prompts can be decomposed into discrete prompt elements \cite{ju2023}. We train a feed-forward neural network to learn how to combine our vectors in $\mathcal{D}$.  Our approach involves training a model on a classification task dataset. Given a set of 4 discrete prompts $\mathcal{D}$ as seen in Table 1, we aim to produce a linear combination of these element's embeddings. Our reasoning in crafting this was that diverse thinking strategies are favorable for mutual orthogonality in the tensor space; our goal was to maximize the span of the tensor space for linear combinations. Below are two different approaches to continuous prompt learning.

\begin{table}[h]
\centering
\begin{tabular}{|>{\raggedright\arraybackslash}p{0.8\linewidth}|}
\hline
Generate a flowchart to visually represent the logic needed to answer the question \\ \hline
Write pseudocode for an algorithm that could determine the answer \\ \hline
Imagine you are explaining the answer to a 5-year-old. Use simple words and analogies. \\ \hline
Summarize the key insights needed to answer in a short poem \\ \hline
\end{tabular}
\caption{First four prompts}
\label{tab:first_four_prompts}
\end{table}

\begin{figure}[h]
    \centering
    \includegraphics[scale=0.7]{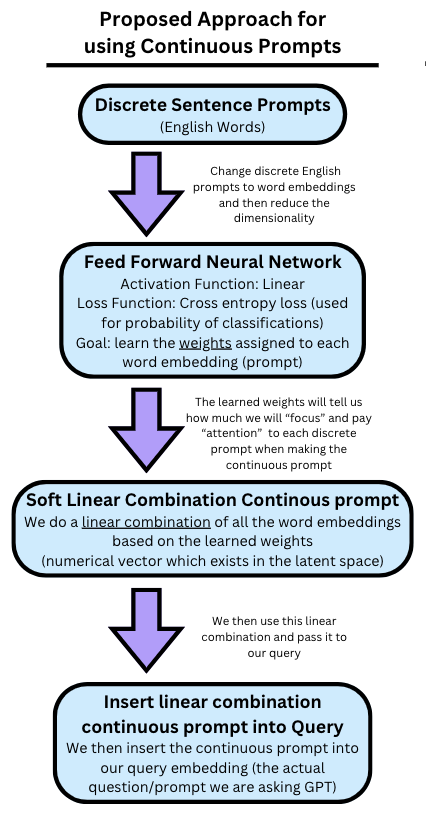}
    \caption{Flow diagram of our Proposed Approach}
    \label{fig:your_label}
    \end{figure}

We first sourced the 'facebook/bart-large' pre-trained language model. For our training dataset we used to 'AI2 Reasoning Challenge (ARC)' containing challenging science questions. We then embedded a basis set of textual discrete prompts and passed them to a feed-forward neural network with three linear layers and three dropout layers. The training was performed on the ARC Challenge dataset using the loss derived from BART. We employed the AdamW optimizer with cross-entropy loss to update the model's weights, allowing us to learn the correct weights correlated to $\mathcal{D}$. We then employed hardware acceleration from an NVIDIA V100 GPU.


\begin{table*}[ht]
\centering
\begin{tabularx}{\textwidth}{|c|X|X|}
\hline
Loss & Question & Top 3 Predicted Contributing Prompts \\
\hline
8.39 & Scientists group animals based on physical features. Trout are classified as fish because of what physical feature? Options: A: Fish have gills. B: Fish eat the same food. C: Fish live in the same area. - D: Fish have the same predators. & I. Generate a flowchart to visually represent the logic needed to answer the question. II. Create a metaphor relating the question to a seemingly unrelated domain. III. Prototype a computer program to compute the answer algorithmically \\
\hline
8.08 &Which would most likely need to happen for a new plant to grow? Options:- A: Leaves grow out of a stem. - B: Insects get attracted to the petals. - C: A blossom falls into the soil.- D: A seed sprouts into a seedling. & I. Generate a flowchart to visually represent the logic needed to answer the question II. Create a metaphor relating the question to a seemingly unrelated domain III. Act out an exaggerated skit to depict the logic behind the answer, \\
\hline
7.93 & A person cuts down a living oak tree. The person burns the wood from the oak tree to boil water. Which sequence correctly orders the energy transformations that occurred from the living tree to the boiling of water? A: light energy → chemical energy → thermal energy - B: thermal energy → chemical energy → light energy - C: chemical energy → mechanical energy → electrical energy - D: electrical energy → mechanical energy → chemical energy & I. Generate a flowchart to visually represent the logic needed to answer the question. II. Act out an exaggerated skit to depict the logic behind the answer. III. Create a metaphor relating the question to a seemingly unrelated domain\\
\hline
\end{tabularx}
\caption{Loss, Question, and Largest Contributing Prompt}
\label{tab:my_table}
\end{table*}

\begin{figure*}[h!]
    \centering
    \includegraphics[scale=.71]{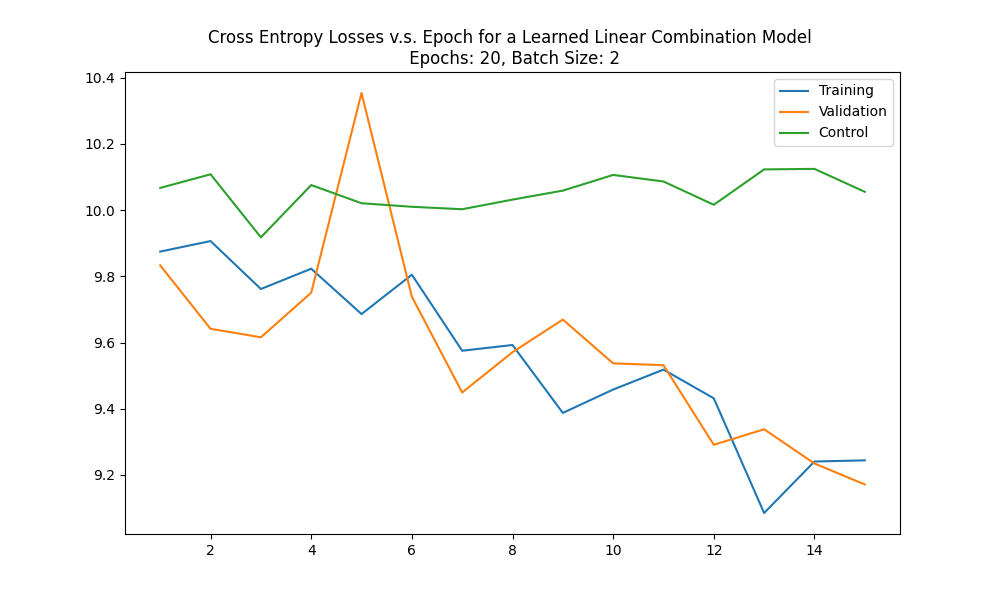}
    \caption{Performance of BART pre-trained model on 20 epochs batch size 2}
    \label{fig:your_label}
\end{figure*}

\begin{figure*}[h!]
    \centering
    \includegraphics[scale=.71]{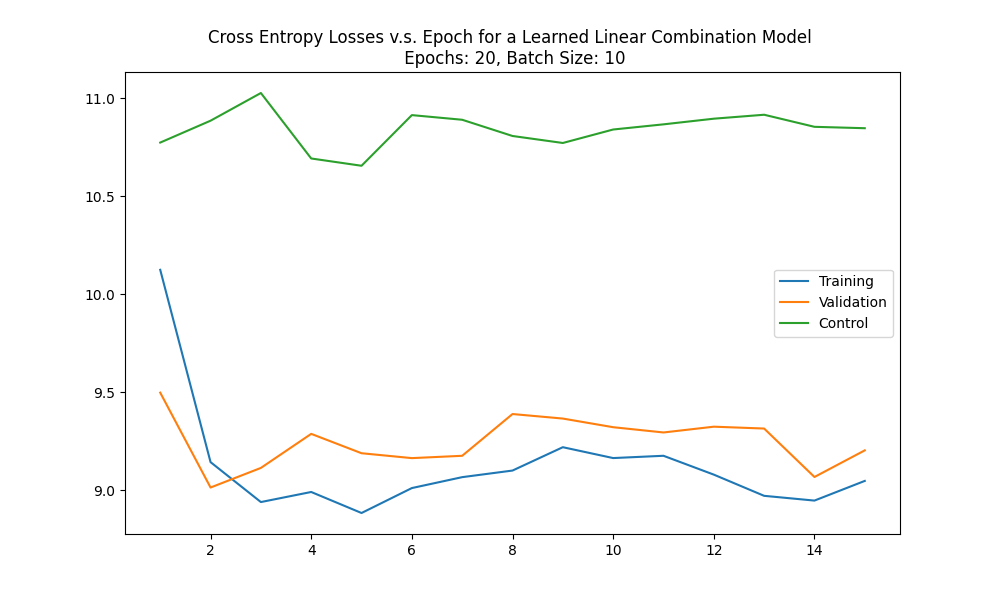}
    \caption{Performance of BART pre-trained model on 20 epochs batch size 10}
    \label{fig:your_label}
\end{figure*}

In analyzing the results we look at the cross entropy loss. Based on this function we perform backpropagation on the feed forward neural network as it learns to predict weights that correspond to loss minimization. In this context we consider the zero tensor as padding. \\

By looking at the predicted weights of the neural model we can compare the magnitude of the weights, and observe which prompts contribute to an observed effect after BART inference. For example if we consider the prompts in Table 1 to be the entirety of $\mathcal{D}$ and our trained neural model predicts a weight-vector $[1.1, 0, -.5, 3]$, geometrically we can make the interpretation that the last prompt contributes most to a favorable performance in a task, as well as an inverse relation for the second to last prompt. 

\section{Experiments and Results}

We ran multiple experiments comparing the performance of our neural model and the control, the control being BART predicting the target sequence without concatenating a continuous prompt embedding to the input of inference.   \\

We ran the training program on both the training and validation splits from ARC. We observed signs of instability with a batch size of 2, but when we increased the batch size to 10 the loss curve stabilized more quickly. We observed that although the loss improved over time in both cases, it's not clear if further training would yield better results. 

We examine the weights assigned by the model for various tasks. Table 2 describes a few examples of the model predicting the weights for use in the linear combination operation. We observed that in particular the prompt saying "generate a flowchart" had the most contribution in the prediction for many of the samples. However, the level of contribution for each from in $\mathcal{D}$ didn't appear to be the same for each question, for example the prompt "Prototype a computer program to compute the answer algorithmically" contributed to a grouping task. 

\subsection{Model Interpretability}
In summary, our architecture is designed to assign scalar weights to each discrete prompt embedding, and them sum them together. Below is an example of one such task:

\begin{table}[H]
    \centering
    \begin{tabular}{|p{3.45cm}|p{3.45cm}|}
        \hline
        \multicolumn{2}{|p{6.6cm}|}{Students are designing kites to discover what type of kite flies the highest. Which is the most important to consider when designing a kite to fly high?} \\
        \hline
         Choice A & string length\\
         \hline
         Choice B & surface area\\
         \hline
         Choice C & materials used\\
         \hline
         Choice D & time of day\\
         \hline
    \end{tabular}
    \caption{Example question A.}
\end{table}

\begin{table}[H]
    \centering
    \begin{tabular}{|p{4.8 cm} | c|}
        \hline
         Basis-Defining Prompt &  Weight \\
         \hline
         "Prototype a computer program to compute the answer algorithmically" & 1.4863\\
         \hline
         "Act out an exaggerated skit to depict the logic behind the answer" & -0.0842 \\
         \hline
         "Summarize the key insights needed to answer in a short poem" & -0.8324\\
         \hline
    \end{tabular}
    \caption{Example basis-defining prompts associated with question A.}
\end{table}

For the task in Table 3, we observe that the most logically motivated prompt had the highest assigned weights - 1.4863. Meanwhile, the prompt concerned with artistic and language focused strategies received less favorable weightings -0.0842 and -0.8324. This behavior is consistent with the assumption that the quality of function of discrete prompts are self-evident.

\section{Conclusion}

We hypothesized that our proposed architecture for the development of continuous prompts via learned linear combinations of discrete prompt embeddings would improve the interpretability of the semantic meanings of generated continuous prompts. Indeed, we observed both a reduction in loss as compared to a control model and weights of basis-defining discrete prompts that, to the general population, would correspond to desired behaviors in the pre-trained language model associated with the dataset. We can thus conclude that restricting the basis of a learned continuous prompt to that which is represented by related discrete prompt embeddings leads to further closure of the cognitive gap between machine shortcuts that arise as a result of the unsupervised learning process and the interpretable, effective knowledge carried by discrete prompts. Improved comprehensibility with loss reduction as defined within the scope of this work is, for the better part, achieved.  

\section{Limitations}
Due to the time and resource constraints we were unable to extend testing to more specialized language understanding tasks. Additionally we hypothesize that performance is a function of the mutual orthogonality of the prompt basis, so an interesting thing to explore would be how different prompt basis would affect the performance.

\bibliography{anthology,custom}
\bibliographystyle{acl_natbib}

\end{document}